\documentclass{article}

\usepackage{arxiv}

\usepackage[utf8]{inputenc} 
\usepackage[T1]{fontenc}    
\usepackage{hyperref}       
\usepackage{url}            
\usepackage{booktabs}       
\usepackage{amsfonts}       
\usepackage{nicefrac}       
\usepackage{microtype}      
\usepackage{lipsum}
\usepackage{fancyhdr}       
\usepackage{graphicx}       
\graphicspath{{media/}}     
\usepackage{makecell}
\usepackage{arydshln}
\usepackage{multirow}
\usepackage[dvipsnames,table]{xcolor}
\usepackage{amsmath}

\usepackage{amssymb}
\newcommand{\bftab}{\fontseries{b}\selectfont}
\usepackage{algorithm}
\usepackage{algpseudocode}

\usepackage{wrapfig}
\usepackage{subcaption}
\def\modelname{FSMisD}

\RequirePackage{xspace}
\makeatletter
\DeclareRobustCommand\onedot{\futurelet\@let@token\@onedot}
\def\@onedot{\ifx\@let@token.\else.\null\fi\xspace}

\def\eg{\emph{e.g}\onedot}
\def\ie{\emph{i.e}\onedot} 
  
\title{Towards Efficient and General-Purpose Few-Shot Misclassification Detection for Vision-Language Models}

\author{
  Fanhu Zeng\textsuperscript{\rm 1, \rm 2}, Zhen Cheng\textsuperscript{\rm 1, \rm 2}, Fei Zhu\textsuperscript{\rm 3}, Xu-Yao Zhang\textsuperscript{\rm 1, \rm 2}\thanks{Corresponding author.} \\
  \textsuperscript{\rm 1}State Key Laboratory of Multimodal Artificial Intelligence Systems, Institute of Automation, CAS\\
  \textsuperscript{\rm 2}School of Artificial Intelligence, UCAS\\
  \textsuperscript{\rm 3}Centre for Artificial Intelligence and Robotics, HKISI-CAS \\
}

\begin{document}

\maketitle

\begin{abstract}
Reliable prediction by classifiers is crucial for their deployment in high security and dynamically changing situations. However, modern neural networks often exhibit overconfidence for misclassified predictions, highlighting the need for confidence estimation to detect errors. Despite the achievements obtained by existing methods on small-scale datasets, they all require training from scratch and there are no efficient and effective misclassification detection~(MisD) methods, hindering practical application towards large-scale and ever-changing datasets. In this paper, we pave the way to exploit vision language model~(VLM) leveraging text information to establish an efficient and general-purpose misclassification detection framework. By harnessing the power of VLM, we construct \textbf{\modelname{}}, a \textbf{F}ew-\textbf{S}hot prompt learning framework for \textbf{MisD} to refrain from training from scratch and therefore improve tuning efficiency. To enhance misclassification detection ability, we use adaptive pseudo sample generation and a novel negative loss to mitigate the issue of overconfidence by pushing category prompts away from pseudo features. We conduct comprehensive experiments with prompt learning methods and validate the generalization ability across various datasets with domain shift. Significant and consistent improvement demonstrates the effectiveness, efficiency and generalizability of our approach.
\end{abstract}

\section{Introduction}
\begin{figure}[t]
    \centering
    \begin{minipage}{0.58\linewidth}
        \includegraphics[width=\textwidth]{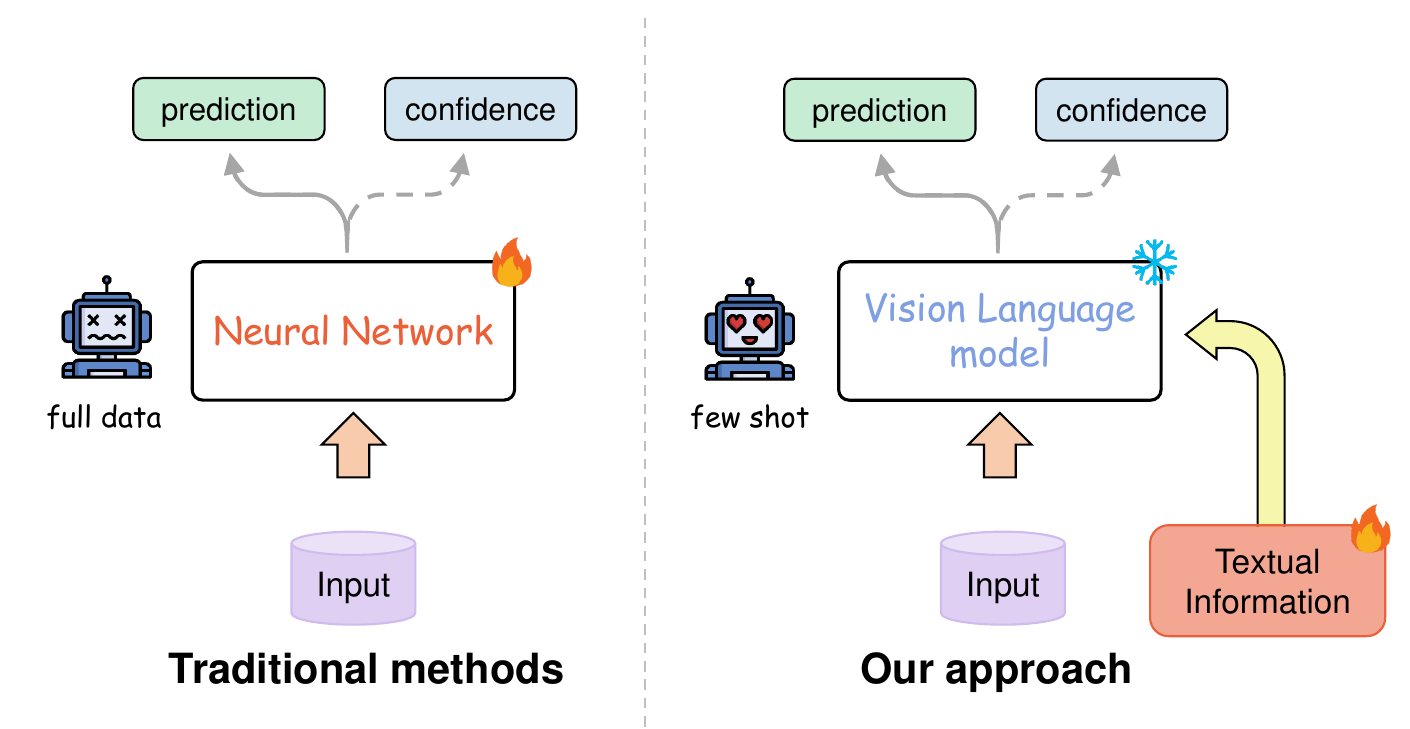} 
    \vspace{-15pt}
    \caption{Overall definition of misclassification detection and comparison between traditional methods and our framework.}
    \label{fig:overall-comparison}
    \end{minipage}
    \hfill
    \begin{minipage}{0.37\linewidth}
        \vspace{-12pt}
        \includegraphics[width=\textwidth]{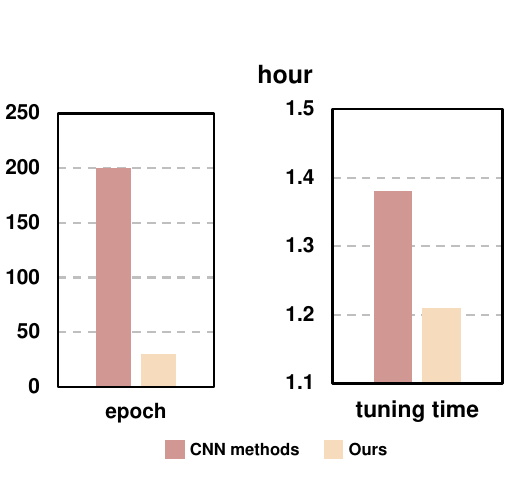}
        \vspace{-20pt}
        \caption{Efficiency comparison between traditional method and the proposed method.}
        \label{fig:efficiency-analysis}
    \end{minipage}
    \vspace{-5pt}
\end{figure}

Neural networks exhibit astonishing performance in diverse tasks and have been deployed in various scenarios. However, casualty is especially crucial for industries that require a high degree of safety, in which wrong predictions are likely to pose security hazards, such as face identification~\cite{geng2020recent}, anomaly detection~\cite{georgescu2021anomaly}, and so on. Unlike humans that inevitably make incorrect predictions and therefore are cautious about their outcomes~\cite{zhu2024open, zhu2022learning}, neural networks are typically trained on close-world assumptions and tend to be overconfident about their predictions, \ie, assign excessively high confidence for erroneous samples~\cite{havasi2020training, hendrycks2016baseline, corbiere2019addressing}. Consequently, it is significant that the classifier is able to classify objects with high confidence while maintaining caution about predictions with low confidence~\cite{cheng2023unified, zhu2024rcl}. In other words, it is for networks to be aware of potential risk in security-sensitive applications and make safe decisions accordingly. The greatest issue lies in the blind confidence in erroneous outcomes~\cite{hendrycks2016baseline}. As a result, Misclassification Detection, the purpose of which is to detect samples with overly high confidence and guarantee security, should receive equal attention along with the prediction ability of the classifiers. As illustrated in Fig.~\ref{fig:overall-comparison}, the goal of MisD is to correctly classify between categories and estimate misclassified samples with low confidence. 

Many efforts have been made to address the issue~\cite{hendrycks2016baseline, zhu2022rethinking, moon2020confidence}. However, existing methods with respect to MisD \textbf{(1)} require training from scratch with full training data~\cite{zhu2023openmix, cheng2023unified, zhu2023revisiting}. \textbf{(2)} Moreover, the ineffectiveness in larger datasets and unacceptable computational complexity~\cite{sun2022out} restrict their evaluation on small-scale datasets including CIFAR~\cite{krizhevsky2009learning}, tiny-ImageNet, and so on. For the former, it is impractical to retrain each model from scratch before deployment, and therefore the ability to defend domain shift is not validated. For the latter, the drawbacks also seriously hinder the application of existing research in practical scenarios, which inherently encompasses diverse amounts of data, and some rare classes may have very few images.

The above two problems naturally raise an open question: \textbf{Can we build a MisD framework that is both time-saving and computationally efficient enough to be extended to large-scale datasets?}

The arrival of vision language model~\cite{radford2021learning} has promoted the research and performance improvement of many downstream tasks~\cite{liu2024visual,xu2021videoclip}. Some parameter-efficient tuning strategies greatly reduce costs while improving efficiency and performance~\cite{jia2022visual,hu2021lora,gao2023clip}. This motivates us to facilitate effective misclassification detection approaches towards large-scale datasets by means of efficiently tuning pre-trained multi-modal representations.

In this paper, we investigate the power of efficient tuning on pre-trained vision language model for MisD task, which has not been well explored before. Specifically, we harness few-shot prompt learning as base framework to train category prompt for each class, therefore \textbf{avoiding training from scratch}, as is illustrated in Fig.~\ref{fig:overall-comparison}. It is notable that the data efficient attribute of our method is \textbf{especially vital when data collection is rather expensive}. We then propose a textual guided negative augmentation to generate adaptive pseudo samples and a novel negative loss to alleviate the overconfidence of the classifier by pushing category prompts away from pseudo features. We conduct comprehensive experiments on large-scale datasets including ImageNet to validate its efficiency. Significant and consistent improvement across different shots and evaluation metrics corroborates the effectiveness of our method. We also carry out experiments on various domain shift datasets to certificate the generalization ability of the approach. Our contributions are summarized as:
\begin{itemize}
    \item To the best of our knowledge, we are the first one to analyze efficient tuning of pre-trained vision language model for misclassification detection.
    \item We build an efficient learning framework for MisD to improve tuning efficiency on large-scale datasets, and propose effective pseudo sample generation along with negative loss to enhance confidence estimation.
    \item We systematically evaluate several prompt learning methods and conduct exhaustive experiments across various datasets to demonstrate the effectiveness, generalizability and superiority of our approach.
\end{itemize}

\section{Related Work}
\subsection{Misclassification Detection.} 
Misclassification detection~\cite{hendrycks2016baseline}, which is also called failure prediction, is intended to distinguish misclassified images by a classifier from corrected ones~\cite{zhu2022learning}. As modern architecture suffers from overconfidence, it is crucial for classifiers to establish reliable predictions in real-world scenarios, where open world does not limit the number of categories and many issues that differ from the closed-world assumption are urgently waiting to be resolved~\cite{guo2024desire,ma2024towards}. In the field of misclassification detection, existing works~\cite{zhu2023openmix,hendrycks2016baseline,cheng2023unified,moon2020confidence} mostly employ deep neural networks~(DNN) to analyze the effect of misclassification detection on small-scale datasets like CIFAR-10~\cite{krizhevsky2009learning} and tiny-ImageNet. Openmix~\cite{zhu2023openmix} exposes outlier samples to help distinguish misclassified images. OVA~\cite{cheng2023unified} treats the classification of $K$ known classes into binary probabilities with a hybrid learning strategy. FMFP~\cite{zhu2022rethinking} significantly reduces the confidence of misclassified samples while maintaining the confidence of correct ones with flat minima. However, they integrate the approach into training procedure to mitigate overconfidence, which limits them into training from scratch and prohibits them from efficiently expanding the scale of data. Unlike previous methods, we efficiently evaluate the performance of MisD on large-scale datasets by means of pre-trained model.

\subsection{Vision Language Models.} 
The emergence of vision language model~(VLMs)~\cite{radford2021learning}, which is unsupervisely pre-trained on large-scale image-text pairs through contrastive learning, has achieved astonishing performance in zero-shot evaluation~\cite{bujwid2021large} and transfer learning~\cite{he2020momentum}. It has greatly promoted a new paradigm for multi-modal representation learning~\cite{dai2023instructblip, li2021align, kim2021vilt,he2022masked}. CLIP~\cite{radford2021learning} introduces a dual-tower paradigm that extracts the feature of image and text individually and obtains stable representations from contrastive learning. InstructBLIP~\cite{dai2023instructblip} directly fuses the multimodal information from the structure of Q-former. In addition to advancing the development of various downstream tasks~\cite{tong2022videomae, guo2025federated, wei2023vary, zeng2025enhancing, guo2025hide}, it also leverages as auxiliary information from other modalities to enhance traditional single-modal tasks including object detection~\cite{chen2022multimodal}, image generation~\cite{rombach2022high}, and so on~\cite{zeng2024modalprompt}. Similarly, utilizing textual information should also be beneficial to misclassification task, which motivates us to leverage the power of vision language model to enhance the performance of misclassification detection.

\subsection{Few-Shot Learning.} 
Few-shot learning of pre-trained models aims to utilize stable representation from large-scale pre-training and efficiently learn knowledge of downstream tasks with scarce samples~\cite{wang2023training}. It is of great significance to transfer knowledge and instantly adapt to task-specific domain~\cite{long2015learning}. Approaches for few-shot learning are mainly divided into two categories: prompt learning and adapter learning. Prompt learning~\cite{zhou2022learning, zhou2022conditional} applies few additional embeddings that are trained end-to-end with data to overcome inefficient computation. Adapter learning adds modules with a small amount of parameters to balance upstream and downstream representations~\cite{gao2023clip,zhang2021tip}, in which LoRA~\cite{hu2021lora} has shown its tremendous potential and gained its popularity in fine-tuning large models~\cite{liu2024visual}. In this paper, we explore efficient few-shot learning algorithm for misclassification detection from pre-trained vision language model and focus on prompt learning due to its simplicity and scalability.

\section{Preliminary and Background}
\subsection{Basic Notations}
Denote $S = \{(\boldsymbol{x}_i,y_i)\}_{i=1}^n$ represents the whole dataset with $n$ samples, where $\boldsymbol{x}_i \in \mathcal{X}$, $y_i \in \mathcal{Y}$ and $\mathcal{X},\mathcal{Y}$ are input and label spaces. Denote the classifier $h$ is the network parameterized with $\theta$ to predict the softmax probability of given sample for each category, \ie, $h_\theta(\boldsymbol{x})$, and allocate the input to the class exhibiting the highest probability, \ie, $\hat{y} = \mathrm{argmax}_{y\in\mathcal{Y}}h_\theta^y(\boldsymbol{x})$.

\subsection{Problem Formulation}

The goal of misclassification detection is to detect misclassified samples from corrected ones based on their confidence rankings.    Meanwhile, the classifier confidence-estimation function $\xi(\cdot)$ is introduced to assess the confidence of the prediction, and a discriminant function $\kappa(\cdot)$ is exploited to distinguish correct samples from misclassified ones:
\begin{equation}
    \kappa(\boldsymbol{x}) =  \left \{
     \begin{aligned}
        & \mathrm{correct},& \qquad \xi(\boldsymbol{x}) \geq \delta \\
        & \mathrm{misclassified},& \qquad \xi(\boldsymbol{x}) < \delta \\ 
     \end{aligned}
    \right.,
\end{equation}
where $\delta$ is the confidence threshold. In practical application, the associated probability $h_\theta^{\hat{y}}(\boldsymbol{x})$, \ie, maximum softmax probability~(MSP), can be viewed as the confidence-estimation function. In few-shot setting, a small part of images of each class~(\eg, $1$, $4$, $8$ or $16$ images) are used for training and the whole validation set is used to certificate the effectiveness of learning knowledge of downstream task efficiently.

\subsection{Feature Extraction of CLIP}
\label{sec:feature-extraction}
CLIP~\cite{radford2021learning} introduces a dual-tower paradigm that exploits separate vision and text encoder to extract multi-modal representations. For a given image $x$ with height $H$ and width $W$, vision encoder~\cite{dosovitskiy2020image} first split the image into non-overlap patches and then extracts visual features from patched images $E_I(\boldsymbol{x})$: $\mathbb{R}^{3 \times H \times W} \rightarrow \mathbb{R}^{d}$, where $d$ is the hidden dimension.

In order to boost single modality tasks involving images, auxiliary textual information can also be exploited to semantically enhance the performance of image-only tasks. For image classification task, a template suffixed with the name of category is employed to supplement the information on the text side. Concretely, the input of text is represented as: $\boldsymbol{\epsilon}_c= [\boldsymbol{v}_1,\cdots,\boldsymbol{v}_L;\boldsymbol{v}_c], c\in \{1,\cdots, C\}$, where $C$ is category number, $\boldsymbol{v}_i$ can be either hand-crafted embeddings or learnable embeddings and $\boldsymbol{v}_c$ is the embedding of category name. Textual information for each class is encoded with text encoder~\cite{vaswani2017attention}: $\boldsymbol{t}_c=E_T(\boldsymbol{\epsilon}_c)$: $\mathbb{R}^{(L+1)\times d} \rightarrow \mathbb{R}^d$ and interacts with image features to promote target task.

\section{Methodology}
\begin{figure}[tb]
    \centering 
    \includegraphics[width=\textwidth]{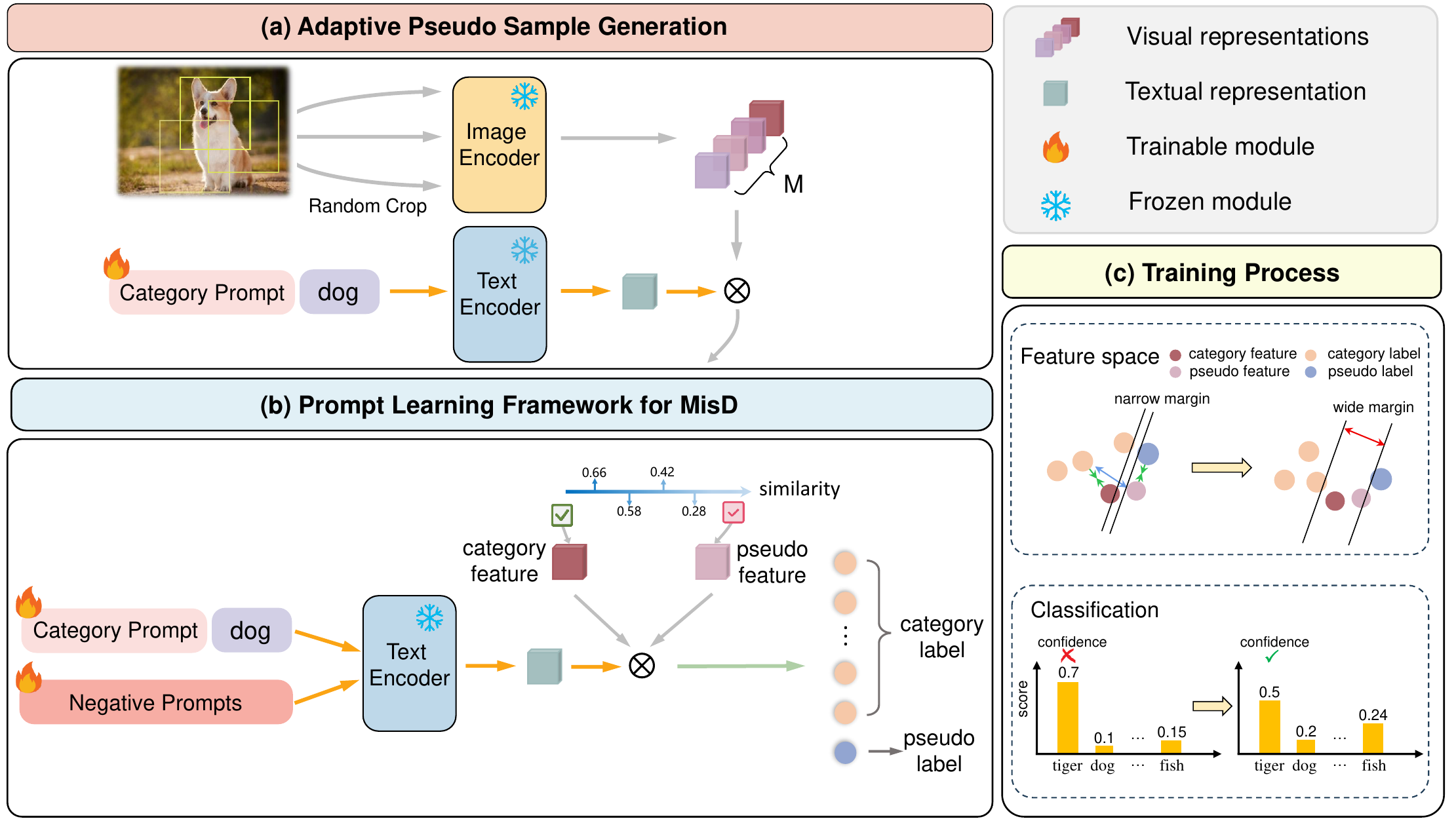}
    \vspace{-10pt}
    \caption{Overall structure of the proposed prompt learning framework for misclassification detection. Adaptive pseudo sample generation is employed to generate pseudo samples and enhance misclassification detection by pushing category prompts away from pseudo labels. Two components are trained together in an end-to-end manner to construct a clear margin for misclassification and enhance confidence estimation.}
    \vspace{-10pt}
    \label{fig:overall-structure}
\end{figure}

In this section, we first implement the efficient prompt learning framework that essencially guarantees the efficiency and is capable of improving MisD with few samples of downstream tasks by means of pre-trained vision language model and then showcase our enhancement to efficiently improve MisD ability. Overall structure is shown in Fig.~\ref{fig:overall-structure}.

\subsection{Category Prompt Construction}
As part of prompt learning in essence, constructing and learning prompts for each category is necessary for efficient few-shot learning. Following existing prompt learning approaches~\cite{zhou2022learning}, we set one learnable category prompt $\boldsymbol{\epsilon}_c$ for each category $c, c\in\{1,\cdots,C\}$ suffixed with the embedding of category name as described in Sec.~\ref{sec:feature-extraction}. It serves as the auxiliary information from textual side to enhance the classification ability. Detailed loss function for classification is introduced in Sec.~\ref{sec:training}.

\subsection{Adaptive Pseudo Sample Generation}
\noindent \textbf{Motivation.} The core and straightforward approach for misclassification detection is to reduce the confidence of the detector in uncertain samples. However, it is hard and time-consuming to manually determine the degree of uncertainty in given image samples. Instead of partitioning the given dataset according to confidence estimation, we propose to generate pseudo samples with low confidence for each image to avoid the complicated and labour-intensive procedure. On the other hand, the pseudo samples can be regarded as more difficult misclassification samples than real ones, and models trained on them can therefore be more discriminative at misclassification detection.

\begin{algorithm}[!ht]
    \caption{Few-Shot Learning Framework for Misclassification Detection}
    \label{alg:training-procedure}
    \begin{algorithmic}[1]
        \Require Input image with label $(\boldsymbol{x},y)$.
        \Ensure Optimized category prompts $\boldsymbol{t}_c$ and negative prompts $\breve{\boldsymbol{t}}$.
        \State $\triangleright$ \begin{sc}Generate pseudo samples and encode features \end{sc} 
        \State \quad \quad $\boldsymbol{\epsilon}_c = [\boldsymbol{v}_1,\cdots,\boldsymbol{v}_L;\mathrm{embedding}(y)]$, \ $\breve{\boldsymbol{\epsilon}} = [\boldsymbol{v}_1,\cdots,\boldsymbol{v}_L]$
        \State \quad\quad $[\boldsymbol{x}_1,\cdots, \boldsymbol{x}_k]$ = RandomCrop($\boldsymbol{x}$)
        \State \quad\quad $\boldsymbol{q}_i$ = $E_I(\boldsymbol{x}_i), i\in\{1,\cdots,k\}$, \ $\boldsymbol{t}_c=E_T(\boldsymbol{\epsilon}_c)$, \ $\breve{\boldsymbol{t}}=E_T(\breve{\boldsymbol{t}})$
        \State \quad\quad $\boldsymbol{q}, \ \hat{\boldsymbol{q}} =\mathrm{\underset{\boldsymbol{q}_i}{argmax}}\ \mathrm{{sim(\boldsymbol{q}_i,\boldsymbol{t}_c)}}, \ \mathrm{\underset{\boldsymbol{q}_i}{argmin}}\ \mathrm{{sim(\boldsymbol{q}_i,\boldsymbol{t}_c)}}$
        \State $\triangleright$ \begin{sc}Training procedure\end{sc}
        \While{training}
        \State \quad  $\triangleright$ Classification restriction on category prompts $\boldsymbol{t}_c$ with $\mathcal{L}_{ce}$ in Eqn.~\eqref{eqn:ce}
        \State \quad  $\triangleright$ Contrastive restriction on negative prompts $\breve{\boldsymbol{t}}$ with $\mathcal{L}_{\mathrm{neg}}$ in Eqn.~\eqref{eqn:neg}
        \State \quad  $\triangleright$ Orthogonal restriction on negative prompts $\breve{\boldsymbol{t}}$ with  $\mathcal{L}_{\mathrm{orth}}$ in Eqn.~\eqref{eqn:orth}
        \State \quad  $\mathcal{L}_{\mathrm{total}} = \mathcal{L}_{\mathrm{ce}} + \lambda_{}  \mathcal{L}_{\mathrm{neg}} + \lambda_{\mathrm{orth}} \mathcal{L}_{\mathrm{orth}}$
        \State \quad  $\triangleright$ Optimize $\boldsymbol{t}_c$ and $\breve{\boldsymbol{t}}$
        \EndWhile
    \end{algorithmic}
\end{algorithm}

For favorable adaptation towards MisD, we practically construct negative samples from augmentation to alleviate overconfidence for misclassified images. The motivation mainly comes from the intuition that certain part of the image with foreground at the center deserves the highest confidence and those with foreground object at the edge or background in the image should be allocated with low confidence. To this end, we exploit augmentation of random cropping that naturally meets the aforementioned demand. We propose textual guided negative augmentation to generate pseudo negative samples for MisD. Specifically, several randomly cropped images of original sample calculate similarity with corresponding textual feature $\boldsymbol{t}_c$ and images with the most similarity are marked as normal samples $\boldsymbol{q}$ for cross-entropy training mentioned above, and that with the least similarity are generated pseudo features $\breve{\boldsymbol{q}}$. Perceptually speaking, the adaptation lies in that a cropped image is labeled as negative sample once it is not an area similar enough to category prompt even if it is essentially part of the input image~(relatively higher overall similarity) and vice versa. We design the following training pipeline to mitigate the over-confidence by ensuring the model to be discriminative between features and pseudo features. The effectiveness of random crop strategy is shown in Sec.~\ref{sec:ablation-study}.

\subsection{Overall Structure for Training}
\label{sec:training}
For training procedure, given a sample pair $(\boldsymbol{x},y)$, $\boldsymbol{x}$ extracts visual category feature $\boldsymbol{q}$ and pseudo feature $\breve{\boldsymbol{q}}$, and textual feature $\boldsymbol{t}_c$ is encoded through $y$. A commonly used cross-entropy loss is first used to classify between categories:
\begin{equation}
\label{eqn:ce}
    \mathcal{L}_{\mathrm{ce}} = \mathrm{crossentropy}(\mathrm{P}(y \vert \boldsymbol{x}), y),
\end{equation}
where the probability is the similarity between image and text features normalized by softmax. It essentially guarantees the effectiveness and efficiency of prompt learning framework~\cite{zhou2022learning}.

Then we use the pseudo features generated above to train for possible misclassification. Concretely, we introduce negative prompts $\breve{\boldsymbol{\epsilon}}$ consisting entirely of learnable context vectors, \ie, $\breve{\boldsymbol{\epsilon}}=[\boldsymbol{v}_1,\cdots,\boldsymbol{v}_L]$ and utilize negative feature $\breve{\boldsymbol{t}}=E_T(\breve{\boldsymbol{\epsilon}})$ to match potential misclassified samples. When pseudo features are introduced, the classifier is encouraged to decrease prediction towards category prompts. To this end, we present a negative loss to mitigate overconfidence by pushing away from category prompts and drawing closer to negative samples. Intuitively, confidence can be decreased about uncertain predictions, \ie, prediction on pseudo features in our framework, when they are intentionally trained to engage with negative prompts. Particularly, negative loss is denoted as:
\begin{equation}
\label{eqn:neg}
    \mathcal{L}_\mathrm{neg}(\breve{\boldsymbol{x}}) =-\mathrm{log}\frac{\sum_{n=1}^{n_\mathrm{n}}\mathrm{exp}(\mathrm{sim}(\breve{\boldsymbol{q}}, \breve{\boldsymbol{t}}_n)/ T)}{\sum_{i=1}^{C}{\mathrm{exp}(\mathrm{sim}(\breve{\boldsymbol{q}}, \boldsymbol{t}_\mathrm{i}) / T)}+\sum_{n=1}^{n_\mathrm{n}}{{\mathrm{exp}(\mathrm{sim}(\breve{\boldsymbol{q}}, \breve{\boldsymbol{t}}_n) / T)}}},
\end{equation}
where $T$ is the temperature and the measurement of similarity is defined as the cosine similarity distance:
\begin{equation}
    \mathrm{sim}(\boldsymbol{\alpha}_i,\boldsymbol{\alpha}_j)=\frac{\boldsymbol{\alpha}_i\cdot\boldsymbol{\alpha}_j}{\left\| {\boldsymbol{\alpha}_i} \right\| \left\| {\boldsymbol{\alpha}_j} \right\|}.
\end{equation}

Through contrastive learning of negative prompts, samples of high confidence are required to be discriminant with pseudo samples of relatively higher confidence and vice versa. The misclassification detection ability is always enhanced with adaptive pseudo sample generation and the effectiveness of the proposed method is thereby certificated.

Additionally, to encourage the diversity of negative prompts, we introduce orthogonalization loss to cover more unseen attributes in feature space and therefore enhance misclassification detection ability, which is defined as: 

\begin{equation}
\label{eqn:orth}
    \mathcal{L}_\mathrm{orth} = \frac{1}{n_\mathrm{n}^2}\sum_{i=1}^{n_\mathrm{n}}\sum_{j=1}^{n_\mathrm{n}}\mathrm{sim}(\breve{\boldsymbol{t}}_i,\breve{\boldsymbol{t}}_j).
\end{equation}

Intuitively, orthogonalized negative prompts minimize the similarity of each pair of negative prompts and are able to represent more combinations of feature space. Therefore, it is helpful to the stability of misclassification detection. The final loss is a weighted sum of the loss mentioned above:
\begin{equation}
    \mathcal{L}_{\mathrm{total}} = \mathcal{L}_{\mathrm{ce}} + \lambda_{}  \mathcal{L}_{\mathrm{neg}} + \lambda_{\mathrm{orth}} \mathcal{L}_{\mathrm{orth}},
\end{equation}
where $\lambda_{\mathrm{n}}$ and $\lambda_{\mathrm{reg}}$ are respective loss coefficients. We derive the whole training framework in detail in Algorithm~\ref{alg:training-procedure}. All prompts are learnable so that the training process can be performed in an end-to-end manner.

\subsection{Misclassification Detection Evaluation}
During evaluation, an extracted image feature from vision encoder calculates similarity with all category prompts, and obtains prediction probability to each category by softmax function~\cite{hendrycks2016baseline}:
\begin{equation}
    \mathrm{P}(y_i\vert \boldsymbol{x}) = \frac{\mathrm{exp}(\mathrm{sim}(\boldsymbol{q},\boldsymbol{t}_i) / T)}{\sum_{c=1}^{C}\mathrm{exp}(\mathrm{sim}(\boldsymbol{q},\boldsymbol{t}_c) /T)}.
\end{equation}

The estimated confidence is defined as the maximum of softmax output:
\begin{equation}
    \xi(\boldsymbol{x}) = \mathrm{max}_{y\in\mathcal{Y}}\mathrm{P}(y_i\vert \boldsymbol{x}),
\end{equation}
and all metrics regarding misclassification detection can be calculated accordingly.

\noindent \textbf{Remarks.} In the designed efficient misclassification detection framework, we do not propose a more dedicated confidence estimation function and take the most commonly employed maximum softmax probability~(MSP) as the confidence estimator for fair comparison. Our contribution for misclassification detection mainly lies in the training procedure, where the introduced pseudo samples adaptively alleviate the overconfidence of categories by pushing pseudo features away from category features by contrastive learning. Consequently,  it simulates the situation of images with over-confidence and significantly reduces misclassification detection when uncertain images occur during inference.

\section{Experimental Results}
\subsection{Experimental Setup}
\noindent \textbf{Implementation details.} For efficient learning, we mainly conduct the experiments on shots of 1, 2, 4, 8 and 16. The length of learnable context vectors is 16. Loss coefficients $\lambda_{\mathrm{neg}}$ and $\lambda_{\mathrm{orth}}$ are 5 and 0.5, respectively. Temperature $T$ is set to be 1 by default. For a fair comparison, all models are trained for 30 epochs along with
$2\times 10^{-3}$ cosine learning rate schedule. Unless otherwise stated, we employ CLIP-VIT-B/16~\cite{radford2021learning} as the backbone to conduct experiments. Owing to the efficiency of few-shot learning with pre-trained vision-language model, we are not restricted to smaller datasets like CIFAR and Tiny-Imagenet, and mainly conduct experiments across diverse ImageNet dataets to showcase the effectiveness and generalization ability of our method. Due to the lack of comparative misclassification detection methods on large-scale datasets for their unbearable computational overhead, we implement two efficient prompt learning based methods CoOp~\cite{zhou2022learning} and LoCoOp~\cite{miyai2023locoop} for comparison. We additionally conduct efficiency analysis with traditional methods on smaller datasets in Sec.~\ref{sec:comp-traidition}.

\noindent \textbf{Evaluation metrics.} Follow existing works~\cite{zhu2023openmix,zhu2022rethinking,hendrycks2016baseline}, we use FPR95~\cite{flach2015precision}, AURC, E-AURC~\cite{geifman2019bias}, AUPR-Success, AUPR-Error~\cite{hendrycks2016baseline}, and AUROC~\cite{davis2006relationship} to evaluate misclassification detection ability. The lower values for FPR-95, AURC, E-AURC and higher values for AUROC, AUPR-Success, AUPR-Error represent better MisD ability. Detailed definitions of each metric can be found in the original papers. Moreover, we also report the accuracy of the validation set to ensure that classification ability is not sacrificed to improve misclassification detection performance.

\begin{table}[t]
    \centering
    \caption{Results of misclassification detection on large-scale ImageNet dataset. We first implement several methods in prompt learning framework and then compare methods in different tuning manners. $\mathbf{Bold}$ represents the best results.}
    \setlength{\tabcolsep}{9pt}
    \vspace{5pt}
        \resizebox{0.95\linewidth}{!}{
            \begin{tabular}{lccccccc}
            \toprule[1.3pt]
             &\textbf{FPR95}$\downarrow$& \textbf{AURC}$\downarrow$ & \textbf{E-AURC}$\downarrow$ & \textbf{AUROC}$\uparrow$ & \textbf{ACC}$\uparrow$&\textbf{AUPR-S}$\uparrow$&\textbf{AUPR-E}$\uparrow$   \\ 
            \midrule
            \emph{Zero-shot} \\
            Baseline~\cite{radford2021learning} &81.52&207.12&144.32&69.69&66.72&80.80&53.38\\
            \midrule
            \emph{Few-shot} &\multicolumn{6}{c}{\hspace{2mm} 1-\emph{shot}} \\
            CoOp~\cite{zhou2022learning} &\bftab{70.53}&124.83&69.32&81.58 &68.59&90.76&64.29\\
            LoCoOp~\cite{miyai2023locoop}&73.53&132.72&75.23&80.51&68.06&89.94&62.65  \\ \rowcolor{gray!20}\textbf{Ours}&70.78&\bftab{121.42}&\bftab{67.40}&\bftab{81.85}&\bftab{68.99}&\bftab{91.06}&\bftab{64.32}\\
            &\multicolumn{6}{c}{\hspace{2mm} 4-\emph{shot}} \\
            CoOp~\cite{zhou2022learning}&70.32&115.93& 64.61& 82.24& 69.72&91.48&64.39\\
            LoCoOp~\cite{miyai2023locoop}&72.28&123.56&71.41&80.79&69.49&90.59&62.29\\
            \rowcolor{gray!20}\textbf{Ours}&\bftab{70.08}&\bftab{112.61}&\bftab{62.59}&\bftab{82.54}&\bftab{70.09}&\bftab{91.78}&\bftab{64.41}\\
             &\multicolumn{6}{c}{\hspace{2mm} 16-\emph{shot}} \\
            CoOp~\cite{zhou2022learning} &69.44&104.04&58.71&82.83& 71.44&92.40&63.45\\
            LoCoOp~\cite{miyai2023locoop}&70.75&109.73&62.99&82.03&71.02&91.82&62.33\\ \rowcolor{gray!20}\textbf{Ours}&\bftab{67.65}&\bftab{100.56}&\bftab{57.07}&\bftab{83.07}&\bftab{72.01}&\bftab{92.65}&\bftab{63.65}\\
            \bottomrule[1.3pt]
        \end{tabular}}
    \vspace{-10pt}
    \label{tab:main-results}
    
\end{table}

\begin{figure}[t]
    \centering 
    \includegraphics[width=1.0\linewidth]{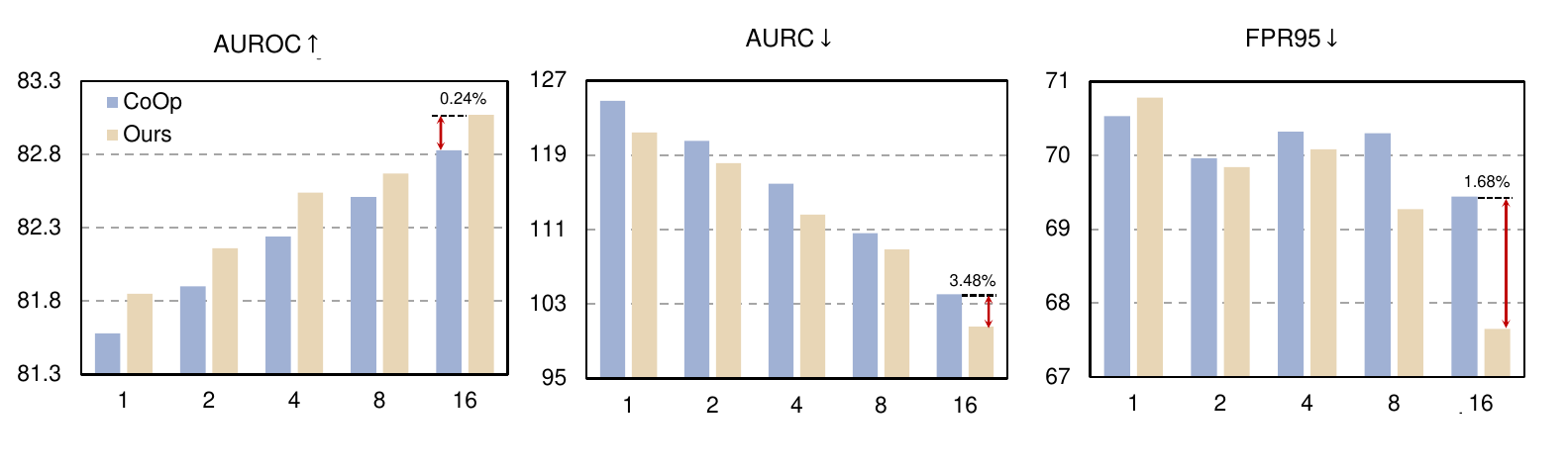}
    \vspace{-30pt}
    \caption{Variation of few-shot misclassification detection performance as samples of each class increase. Outcomes of AURC, AUROC, and FPR95 are reported, respectively.}
    \vspace{-15pt}
    \label{fig:main-shots}
\end{figure}

\subsection{Main Results on ImageNet}
We use ImageNet~\cite{deng2009imagenet}, which collects 1 million images from 1000 categories along with a 100-class subset of ImageNet named ImageNet-100 as datasets to evaluate the performance of misclassification detection on large-scale dataset.

\noindent \textbf{Our method is efficient and effective in large-scale ImageNet-1k.} Firstly, we systematically evaluate several prompt learning methods on misclassification tasks in our framework and outcomes are shown in Tab.~\ref{tab:main-results}. We then conduct experiments with our approach and the results are shown in \textcolor{gray}{gray} lines. It can be found that LoCoOp, which is a prompt learning method designed for open set classification, does not obtain better performance than baseline CoOp, showcasing the difficulty in promoting misclassification detection ability under few-shot setting. By contrast, it is evident that our approach acquires remarkable improvements with respect to all metrics. For example, we obtain notably 1.68\% lower FPR95($\downarrow$) and 0.24\% higher AUROC($\uparrow$) with 16 shots, strongly demonstrating the effectiveness of our method.

\begin{table}[t]
    \centering
    \caption{Experiments on ImageNet-100 dataset with semantic similarity. The best results are highlighted in \textbf{bold}, the second best results are underlined.}
    \setlength{\tabcolsep}{8pt}
    \vspace{5pt}
    \label{tab:imagenet100}
        \resizebox{0.95\linewidth}{!}{
            \begin{tabular}{lccccccc}
            \toprule[1.3pt]
             &\textbf{FPR95}$\downarrow$& \textbf{AURC}$\downarrow$ & \textbf{E-AURC}$\downarrow$ & \textbf{AUROC}$\uparrow$ & \textbf{ACC}$\uparrow$&\textbf{AUPR-S}$\uparrow$&\textbf{AUPR-E}$\uparrow$   \\ 
            \midrule
            Baseline~\cite{radford2021learning}&66.45& 45.99& 38.04&78.71&87.66&95.79&41.40\\
            CoOp~\cite{zhou2022learning}&\underline{58.15}& \underline{17.94}&\bftab{12.31}&\underline{89.40}&\underline{89.58}&\underline{98.64}&47.42\\
            LoCoOp~\cite{miyai2023locoop}&66.31&36.87&24.89&86.13& 84.92&97.18&\underline{50.50}\\
            \rowcolor{gray!20} \textbf{Ours}&\bftab{54.74}&\bftab{17.65}&\underline{12.35}&\bftab{89.73}&\bftab{89.88}&\bftab{98.66}&\bftab{50.63}\\
            \bottomrule[1.3pt]
        \end{tabular}}
    \vspace{-10pt}
\end{table}

\begin{table}[t]
    \centering
    \caption{Misclassification detection performance of different variants. We report 16-shot results of different methods for comparison.}
    \setlength{\tabcolsep}{8pt}
    \vspace{5pt}
    \label{tab:variant}
        \resizebox{0.95\linewidth}{!}{
            \begin{tabular}{cl ccccccc}
            \toprule[1.3pt]
            \textbf{Network}&\textbf{Method}&\textbf{FPR95}$\downarrow$& \textbf{AURC}$\downarrow$ & \textbf{E-AURC}$\downarrow$ & \textbf{AUROC}$\uparrow$ & \textbf{ACC}$\uparrow$&\textbf{AUPR-S}$\uparrow$&\textbf{AUPR-E}$\uparrow$   \\ 
            
            \midrule
            \multirow{3}{*}{ViT-B-16}&CoOp~\cite{zhou2022learning} &69.44&104.04&58.71&82.83& 71.44&92.40&63.45\\
            &LoCoOp~\cite{miyai2023locoop}&70.75&109.73&62.99&82.03&71.02&91.82&62.33\\ 
            &\cellcolor{gray!20}\textbf{Ours}&\cellcolor{gray!20}\bftab{67.65}&\cellcolor{gray!20}\bftab{100.56}&\cellcolor{gray!20}\bftab{57.07}&\cellcolor{gray!20}\bftab{83.07}&\cellcolor{gray!20}\bftab{72.01}&\cellcolor{gray!20}\bftab{92.65}&\cellcolor{gray!20}\bftab{63.65}\\
            \midrule
            \multirow{3}{*}{ResNet-50}&CoOp~\cite{zhou2022learning}&72.06&155.36&78.19&81.77&63.38&88.93&70.81\\
            &LoCoOp~\cite{miyai2023locoop}&72.40&160.59&81.47&81.31&62.95&88.42&69.53\\
            &\cellcolor{gray!20}\textbf{Ours}&\cellcolor{gray!20}\bftab{71.44}&\cellcolor{gray!20}\bftab{152.17}&\cellcolor{gray!20}\bftab{76.49}&\cellcolor{gray!20}\bftab{81.89}&\cellcolor{gray!20}\bftab{65.71}&\cellcolor{gray!20}\bftab{89.23}&\cellcolor{gray!20}\bftab{71.58}\\
            \bottomrule[1.3pt]
        \end{tabular}}
    \vspace{-10pt}
\end{table}

\noindent \textbf{Improving misclassification detection is difficult with increased shots.} We also showcase variation trends about the three most important metrics in MisD, namely AUROC, AURC, and FPR95, under different shots in Fig.~\ref{fig:main-shots}. It can be concluded that the performance gradually gets better with marginal benefits as the number of shots increases, \ie, more shots contribute slightly to the performance improvement, showcasing the difficulty in enhancing misclassification detection ability. By contrast, our approach gets a consistent lead against previous methods. More importantly, it is notable that our method even outperforms more shot results from previous methods under certain circumstances, \eg, ours 4 shot~(82.54\%) against CoOp 8 shot~(82.51\%) in AUROC~($\uparrow$) and ours 2 shot~(69.84\%) against CoOp 8 shot~(70.30\%) in FPR95~($\downarrow$). Considering that the increase gained from more shots is diminishing, the improvement of our method is substantial.
            
\noindent \textbf{Our method is effective on ImageNet-100 with similar semantics.} We select a 100-class subset of ImageNet and assess our approach on smaller dataset with similar semantics. It can be seen in Tab.~\ref{tab:imagenet100} that the proposed approach consistently surpasses existing methods. Moreover, the accuracy~(approximately up to $90\%$) and all MisD metrics are fairly good although the diversity of ImageNet-100 is richer than the datasets used in previous methods, which can be seen in Tab.~\ref{tab:cifar100}. It further validates the superiority of pre-trained VLMs on MisD tasks.

\begin{figure}[tb]
    \centering 
    \includegraphics[width=0.9\textwidth]{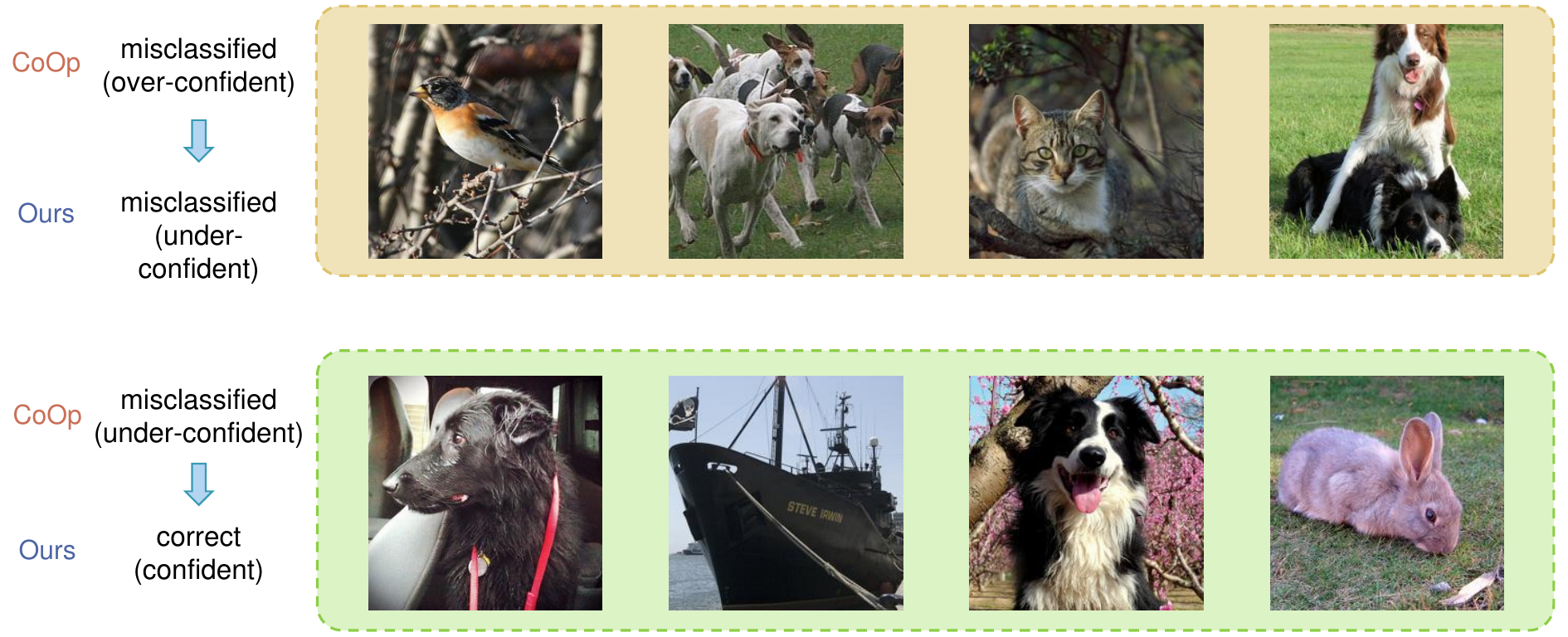}
    \vspace{-5pt}
    \caption{Visualization of different cases in MisD. Compared with previous method, our method successfully mitigates overconfidence and improves MisD performance.}
    \vspace{-10pt}
    \label{fig:visualization}
\end{figure}

\noindent \textbf{Our method is robust to different variants.} Since our method relies on random cropping to generate pseudo samples, it is meaningful to explore the impact of different visual backbones for visual extraction. Different from transformer-based visual encoder that does not alter the receptive field of images, it is also significant to validate the influence of CNN-based backbone, which gradually increases the receptive field. Consequently, We also use Resnet50 as image encoder backbone to validate the generalizability of the proposed method. It is observed in Tab.~\ref{tab:variant} that the model with CLIP backbone receives better performance, which could be attributed to the characteristic of spatial invariance that captures fine information from random cropping. The results also indicate that our method obtains consistent improvements against existing methods with respect to all metrics and all shots, further proving the effectiveness.

\noindent \textbf{Visualization.} 
We additionally provide some visualization results to intuitively illustrate the effectiveness of our method. We showcase typical detection results of CoOp and our method in Fig.~\ref{fig:visualization}. We discover that our model excels at diminishing confidence when the background is complicated or there are multiple objects in the scene, as is depicted on the top of Fig~\ref{fig:visualization}. Conversely, when the object is clear, free from interference, and centrally positioned in the foreground, the proposed model can enhance prediction accuracy with greater confidence, as shown at the bottom of Fig.~\ref{fig:visualization}. It can be summarized that our method not only successfully decreases the confidence of misclassified samples but also improves the confidence of correct samples. Therefore, the utility of the proposed approach is verified explicitly.

\begin{table}[t]
    \centering
    \caption{Generalization ability of misclassification detection methods under natural adversary. Performance of 4-shot and 16-shot are reported for comparison.}
    \vspace{5pt}
    \setlength{\tabcolsep}{5pt}
    \label{tab:natural-adversary}
        \resizebox{\linewidth}{!}{
            \renewcommand{\arraystretch}{1.25}
            \begin{tabular}{l cccccc cccccc}
            \toprule[1.3pt]
            \multirow{2}{*}{\bftab{Method}} &\multicolumn{6}{c}{\bftab{ImageNet-A}} &\multicolumn{6}{c}{\bftab{ImageNet-O}}\\
            \cmidrule(lr){2-7} \cmidrule(lr){8-13} 
             &\textbf{FPR95}$\downarrow$& \textbf{AURC}$\downarrow$ & \textbf{E-AURC}$\downarrow$ & \textbf{AUROC}$\uparrow$ &\textbf{AUPR-S}$\uparrow$&\textbf{AUPR-E}$\uparrow$& \textbf{FPR95}$\downarrow$& \textbf{AURC}$\downarrow$ & \textbf{E-AURC}$\downarrow$ & \textbf{AUROC}$\uparrow$ &\textbf{AUPR-S}$\uparrow$&\textbf{AUPR-E}$\uparrow$\\  
            \midrule
            \emph{Zero-shot} \\
            Baseline~\cite{radford2021learning}&70.29&356.23&149.37&72.54&72.16&73.26&86.26&445.13&179.00&74.41&62.84&72.10\\
            \midrule
            \emph{Few-shot}&\multicolumn{12}{c}{\hspace{2mm} \large{4-\emph{shot}}} \\
            CoOp~\cite{zhou2022learning}&77.66&293.83&139.31&76.50&76.78&74.46&78.80&393.46&161.93&75.49&68.33&79.52\\
            LoCoOp~\cite{miyai2023locoop}&\bftab{76.59}&295.59&145.17&76.19&75.61&74.67&79.33&371.33&\bftab{154.40}&75.76&70.58&79.89\\

            \rowcolor{gray!20} \textbf{Ours}&76.71&\bftab{289.48}&\bftab{137.04}&\bftab{76.71}&\bftab{76.86}&\bftab{75.61}&\bftab{75.03}&\bftab{365.76}&155.74&\bftab{76.14}&\bftab{71.42}&\bftab{81.69}\\

            &\multicolumn{12}{c}{\hspace{2mm} \large{16-\emph{shot}}} \\
            CoOp~\cite{zhou2022learning}&75.65&284.30&136.55&77.21&77.12&76.54&78.39&371.09&155.17&76.21&70.52&81.05\\
            LoCoOp~\cite{miyai2023locoop}&74.48&287.35&142.96&76.93&76.73&75.42&74.89&356.59&150.76&76.37&72.16&82.26\\

            \rowcolor{gray!20} \textbf{Ours}&\bftab{74.38}&\bftab{280.36}&\bftab{132.07}&\bftab{77.38}&\bftab{77.64}&\bftab{77.67}&\bftab{72.39}&\bftab{350.57}&\bftab{148.32}&\bftab{76.21}&\bftab{73.46}&\bftab{82.66}\\
            
            \bottomrule[1.3pt]
        \end{tabular}}
        \vspace{-5pt}
\end{table}

\begin{figure}[ht]
    \centering
    \includegraphics[width=\linewidth]{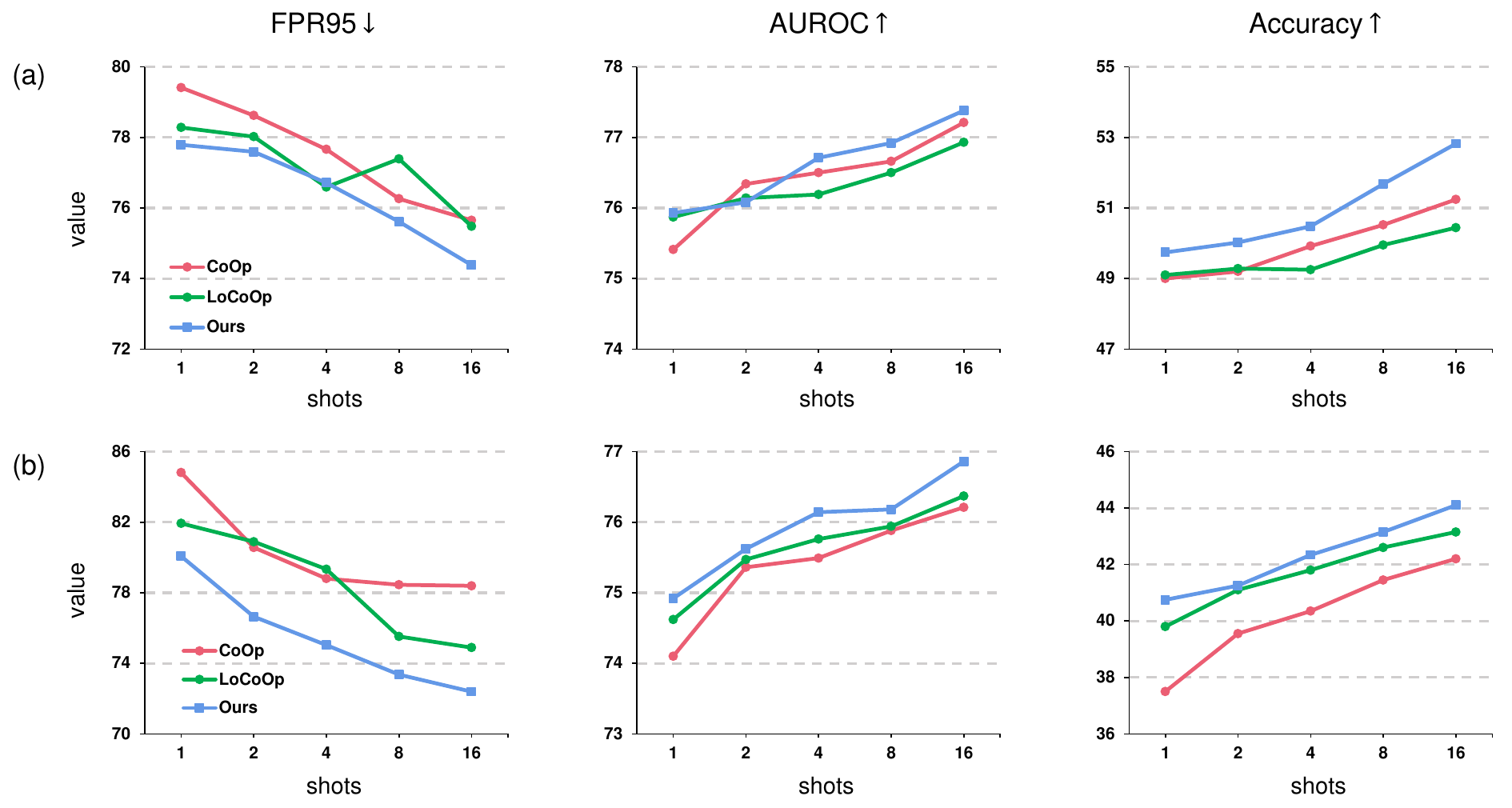}
    \vspace{-15pt}
    \caption{Misclassification detection and classification ability under natural adversary with different shots of each category on (a) ImageNet-A and (b) ImageNet-O. }
    \label{fig:natural-shots}
    \vspace{-10pt}
\end{figure}

\subsection{Misclassification detection under natural adversary}
Owing to the efficiency, it is also significant to evaluate the robustness of misclassification detection algorithm. For example, if a facial recognition system fails to recognize a deliberately dressed person, it is likely to pose a major challenge to security. Therefore, it is vital for the detector to distinguish real-world, unmodified, and naturally occurring examples for the safety of risk-sensitive scenarios and make reliable predictions instantly.

\noindent \textbf{Experimental results.} We utilize ImageNet-A~\cite{hendrycks2021natural} and ImageNet-O~\cite{hendrycks2021natural} to evaluate the performance of misclassification detection against natural adversary. Concretely, they contain 200 categories of ImageNet that are visually distinct from the images in the original ImageNet, focusing on categories that may not share common attributes with ImageNet classes. Therefore, they evaluate how well models trained on ImageNet can generalize to objects or categories that are not part of the original training data.

We conduct experiments on $1$,$2$,$4$,$8$,$16$ shots and commpare with CoOp~\cite{zhou2022learning} and LoCoOp~\cite{miyai2023locoop}. The results in Tab.~\ref{tab:natural-adversary} illuminates that all misclassification detection methods undergo severe performance drop under natural adversary. The phenomenon could be primarily attributed to that the pre-trained model is vulnerable to adversarial attacks, and the incremental gain from increased shots of images is approximately the same as other datasets, certificating the effectiveness of the proposed \modelname{}. For instance, from 1 shot to 16 shot, our method improves AUROC~($\uparrow$) by 1.94\% on ImageNet-O and 1.22\% on ImageNet-1k. It validates the effectiveness of few-shot learning for misclassification detection under natural adversary. It is also worth emphasizing that while existing prompt learning methods can not obtain outstanding performance on all natural adversary datasets, our method still obtains consistently better detection ability with higher classification accuracy. As illustrated in Tab.~\ref{tab:natural-adversary} and Fig.~\ref{fig:natural-shots}, our approach gets obvious improvement 3.77\% and 2.50\% lower FPR95~($\downarrow$) on 4 shot and 16 shot of ImageNet-O, respectively, demonstrating its effectiveness and efficiency under natural adversary.

\begin{table}[t]
    \centering
    \caption{Generalization ability of misclassification detection methods under domain transfer. Performance of 4-shot and 16-shot are reported for comparison.}
    \vspace{5pt}
    \setlength{\tabcolsep}{5pt}
    \label{tab:domain-transfer}
        \resizebox{\linewidth}{!}{
            \renewcommand{\arraystretch}{1.25}
            \begin{tabular}{l cccccc cccccc}
            \toprule[1.3pt]
            \multirow{2}{*}{\bftab{Method}} &\multicolumn{6}{c}{\bftab{ImageNet-R}} &\multicolumn{6}{c}{\bftab{ImageNet-Sketch}}\\
            \cmidrule(lr){2-7} \cmidrule(lr){8-13} 
             &\textbf{FPR95}$\downarrow$& \textbf{AURC}$\downarrow$ & \textbf{E-AURC}$\downarrow$ & \textbf{AUROC}$\uparrow$ &\textbf{AUPR-S}$\uparrow$&\textbf{AUPR-E}$\uparrow$& \textbf{FPR95}$\downarrow$& \textbf{AURC}$\downarrow$ & \textbf{E-AURC}$\downarrow$ & \textbf{AUROC}$\uparrow$ &\textbf{AUPR-S}$\uparrow$&\textbf{AUPR-E}$\uparrow$\\  
            \midrule
            \emph{Zero-shot} \\
            Baseline~\cite{radford2021learning}&60.25&84.22&41.80&87.04&94.60&65.03&74.82&188.39&97.15&79.40&85.81&60.82\\
            \midrule
            \emph{Few-shot}&\multicolumn{12}{c}{\hspace{2mm} \large{4-\emph{shot}}} \\
            CoOp~\cite{zhou2022learning}&58.33&69.11&35.12&87.96&95.57&68.61&69.89&115.31&64.36&82.17&91.23&64.08\\
            LoCoOp~\cite{miyai2023locoop}&58.26&\bftab{65.20}&\bftab{34.02}&87.67&95.64&66.93&73.91&124.47&72.16&80.59&90.49&62.39\\

            \rowcolor{gray!20} \textbf{Ours}&\bftab{56.94}&65.32&34.46&\bftab{88.33}&\bftab{95.66}&\bftab{69.92}&\bftab{69.51}&\bftab{113.47}&\bftab{63.00}&\bftab{82.41}&\bftab{91.72}&\bftab{64.51}\\

            &\multicolumn{12}{c}{\hspace{2mm} \large{16-\emph{shot}}} \\
            CoOp~\cite{zhou2022learning}&57.80&65.16&33.07&88.27&95.89&\bftab{70.50}&68.38&104.57&58.91&82.88&92.36&64.96\\
            LoCoOp~\cite{miyai2023locoop}&57.30&64.27&32.94&87.90&95.94&68.01&70.27&109.81&63.02&81.97&91.82&62.68\\

            \rowcolor{gray!20} \textbf{Ours}&\bftab{56.34}&\bftab{62.62}&\bftab{32.34}&\bftab{88.41}&\bftab{95.99}&70.16&\bftab{67.67}&\bftab{100.89}&\bftab{57.18}&\bftab{84.03}&\bftab{94.03}&\bftab{65.80}	\\
            
            \bottomrule[1.3pt]
        \end{tabular}}
\end{table}

\subsection{Misclassification detection under domain transfer}
Domain shift commonly occurs under diverse and ever-changing real-world scenarios. For example, the detector should be able to discriminate images with similar semantics of different styles under different scenes like cartoon and sketch. To this end, ImageNet-R~\cite{hendrycks2021many} and ImageNet-Sketch~\cite{wang2019learning} are exploited to substantiate misclassification detection ability under domain transfer. They consist of 200/1000 categories of ImageNet designed to evaluate model robustness against various rendering transformations/hand-drawn sketches. It challenges models to recognize objects under altered visual conditions, going beyond standard photographic images and is used to assess generalization and performance on distorted or stylized images.

\begin{figure}[t]
    \centering
    \includegraphics[width=\linewidth]{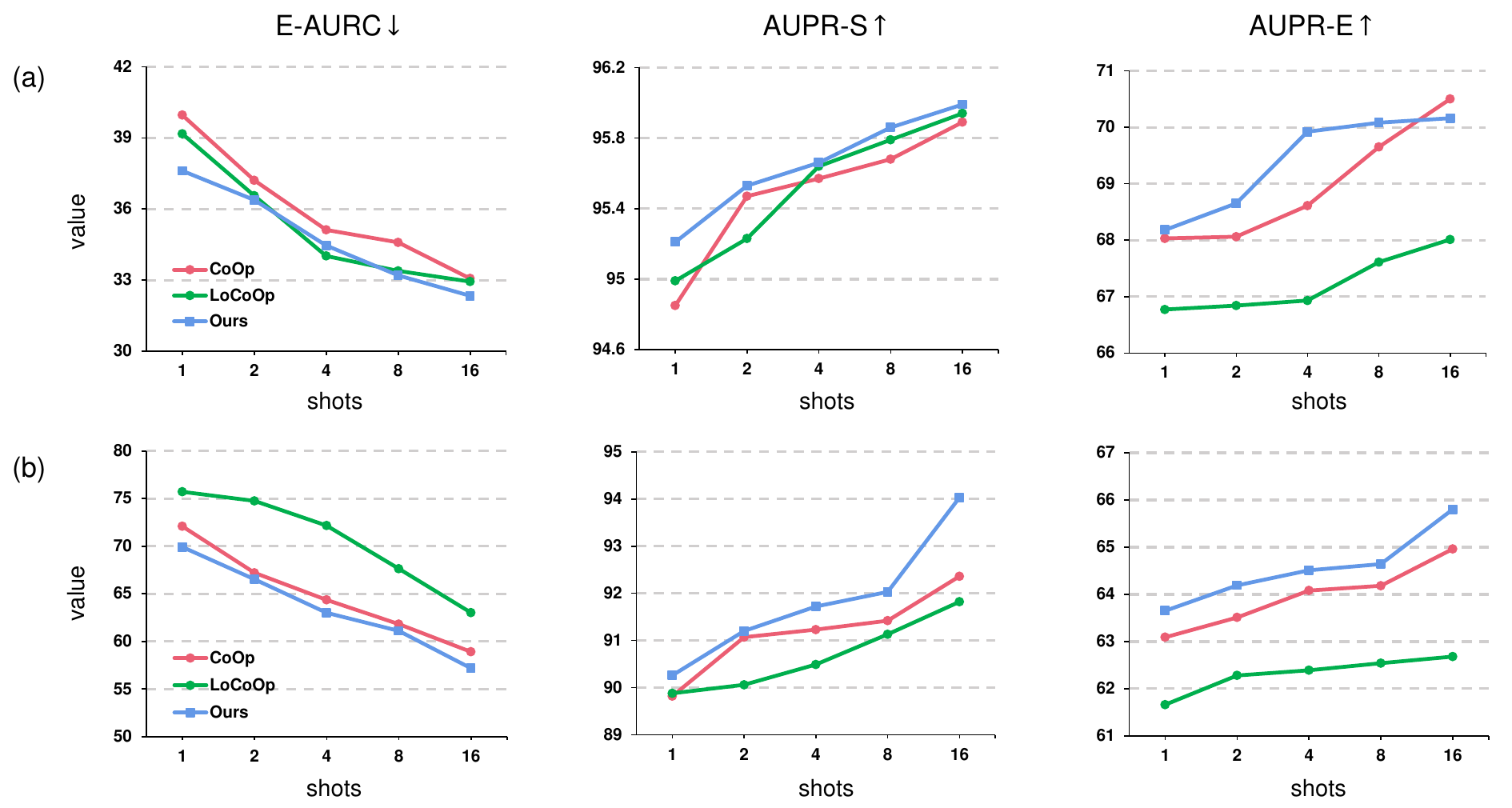}
    \vspace{-15pt}
    \caption{Misclassification detection ability under domain transfer measured by E-AURC, AUPR-S and AUPR-E with different shots of each category on (a) ImageNet-R and (b) ImageNet-Sketch.}
    \label{fig:transfer-shots}
    \vspace{-10pt}
\end{figure}

\noindent \textbf{Experimental results.} Outcomes in Tab.~\ref{tab:domain-transfer} showcase that few-shot learning methods generally achieve better misclassification detection performance than natural adversary, certificating the utility of pre-trained models under domain transfer. Still, our method is able to gain notable improvements with respect to all metrics across both datasets, exhibiting the generalization ability of the proposed method. For example, we achieve  0.71\% lower FPR95~($\downarrow$) and 1.15\% higher AUROC~($\uparrow$) on ImageNet-Sketch. Variation of different shots in Fig.~\ref{fig:transfer-shots} also shows that our method achieves consistent and substantial promotion on all shots without sacrificing classification accuracy. It is worth highlighting that our approach is capable of achieving competitive results with fewer shots, \eg, ours 8 shots against 16 shot LoCoOp on ImageNet-R, which is extremely significant with diminishing marginal benefits. All the results above strongly demonstrate that our approach is beneficial for enhancing misclassification ability under domain transfer.

\subsection{Ablation study}
\label{sec:ablation-study}
\noindent \textbf{Effectiveness of each loss component.} As stated in Sec.~\ref{sec:training}, all loss components have their unique functions intuitively and experimentally. We gradually extend loss components to verify the effectiveness of each proposed loss and report the results in Tab.~\ref{tab:loss-function-component}. It implies that the proposed negative loss is beneficial to alleviate the overconfidence and improve the performance of misclassification detection by forcing pseudo features towards pseudo prompts. Moreover, orthogonalization loss further enhances the stability and contributes to the expression of the classifier. Consequently, we use all of them in the main experiments.

\noindent \textbf{Random cropping is vital for distinguishing misclassified samples.} We alter the way of generating pseudo samples to figure out the impact on distinguishing correct and misclassified samples. The outcomes in Tab.~\ref{tab:augmentation-strategy} indicate that random crop achieves the best performance. Moreover, inappropriate augmentation may lead to counterproductive effects. We attribute the improvement of misclassification detection to the spatial-invariance of random crop that the only uncertainty comes from the displacement of objects other than any introduced noise like gaussian noise, cutout and so on.

\begin{table}[t]
    \centering
    \caption{Effectiveness of each loss component. Baseline refers to employing normal cross-entropy loss only.}
    \vspace{5pt}
    \setlength{\tabcolsep}{8pt}
    \label{tab:loss-function-component}
    \resizebox{\linewidth}{!}{
            \renewcommand{\arraystretch}{1.25}
            \begin{tabular}{lccccccc}
            \toprule[1.3pt]
             &\textbf{FPR95}$\downarrow$& \textbf{AURC}$\downarrow$ & \textbf{E-AURC}$\downarrow$ & \textbf{AUROC}$\uparrow$ & \textbf{ACC}$\uparrow$&\textbf{AUPR-S}$\uparrow$&\textbf{AUPR-E}$\uparrow$   \\ 
            \hline
            Baseline~($\mathcal{L}_\mathrm{ce}$)&70.25&115.17&63.79&82.41&69.71&91.59&64.64\\
            
            $+\mathcal{L}_\mathrm{n}$&70.37&115.76&64.47&82.25&69.73&91.51&64.37\\
            $+\mathcal{L}_\mathrm{n}+\mathcal{L}_\mathrm{o}$&70.08&112.61&62.59&82.54&70.09&91.78&64.41\\
            \bottomrule[1.3pt]
        \end{tabular}}
    \vspace{-5pt}
\end{table}

\begin{table}[t]
    \centering
    \caption{Influence of different augmentation strategies that generate pseudo samples.}   
    \vspace{5pt}
    \setlength{\tabcolsep}{8pt}
        \label{tab:augmentation-strategy}
        \resizebox{\linewidth}{!}{
            \begin{tabular}{lccccccc}
            \toprule[1.3pt]
             \textbf{Component}&\textbf{FPR95}$\downarrow$& \textbf{AURC}$\downarrow$ & \textbf{E-AURC}$\downarrow$ & \textbf{AUROC}$\uparrow$ & \textbf{ACC}$\uparrow$&\textbf{AUPR-S}$\uparrow$&\textbf{AUPR-E}$\uparrow$   \\ 
            \midrule
            Baseline\\
            \quad +Cutout&\bftab{70.01}&116.40&64.95&82.18& 69.69&91.44&64.12\\
            \quad +Gaussian Noise &71.48&118.26&66.13&81.90&69.50&91.27&63.91\\
            \quad \bftab{+Random Crop}&70.08&\bftab{112.61}&\bftab{62.59}&\bftab{82.54}&\bftab{70.09}&\bftab{91.78}&\bftab{64.41}\\
            \bottomrule[1.3pt]
        \end{tabular}}
    \vspace{-10pt}
\end{table}

\noindent \textbf{Focusing on overall information better helps misclassification.} To further validate the usefulness of random cropping, we additionally alter negative loss by pushing away all local features away from category prompt. Results in~\ref{tab:negative-strategy} show that no obvious promotion is observed. One possible reason for the phenomenon would be that random cropping leverages strong local information. In other words, by selecting different regions of the image, the detector gains knowledge of different parts of objects, \eg, foreground, background and edge. Therefore the detector is spatial-aware without the demand of additional local information from feature extraction. Considering both computational efficiency and the performance, we merely employ global information in \modelname{}.

\begin{table}[t]
    \centering
    \caption{Influence of different negative optimization methods.}    
    \vspace{5pt}
    \setlength{\tabcolsep}{8pt}
    \label{tab:negative-strategy}
    \begin{small}
        \resizebox{\linewidth}{!}{
            \begin{tabular}{lccccccc}
            \toprule[1.3pt]
             &\textbf{FPR95}$\downarrow$& \textbf{AURC}$\downarrow$ & \textbf{E-AURC}$\downarrow$ & \textbf{AUROC}$\uparrow$ & \textbf{ACC}$\uparrow$&\textbf{AUPR-S}$\uparrow$&\textbf{AUPR-E}$\uparrow$   \\  
            \midrule
            
            Global&70.08&112.61&62.59&82.54&70.09&91.78&64.41\\
            Local&70.36&115.28&64.38&82.22&69.84&91.53&64.40\\
            Global$+$Local &69.74&114.27&63.57&82.40&69.90&91.63&64.43\\
            \bottomrule[1.3pt]
        \end{tabular}}
        \vspace{-10pt}
    \end{small}
\end{table}

\subsection{Comparison with traditional methods}
\label{sec:comp-traidition}
We compare our pre-trained VLM based few-shot learning method with several traditional misclassification detection algorithms that are trained from scratch including GradNorm~\cite{huang2021importance}, MaxLogit~\cite{hendrycks2022scaling}, CRL~\cite{moon2020confidence} and FMFP~\cite{zhu2022rethinking} to have a clear comparison of the two types of MisD paradigm.

\noindent \textbf{Few-shot learning also contributes to smaller datasets.} For the purpose of achieving an intuitive understanding and facilitating comprehensive comparison with conventional full training methods, and due to the difficulty in implementing existing methods on large-scale datasets, we additionally attest to smaller database CIFAR-10/CIFAR-100~\cite{krizhevsky2009learning} that are used by traditional methods. Results in Tab.~\ref{tab:cifar100} elucidate that we also achieve competitive results when database becomes small, indicating the generalizability of our approach. We underscore that due to the mismatch of resolution, smaller datasets like CIFAR-10 and CIFAR-100 are not commonly evaluated by VLM in similar tasks like out-of-distribution detection~\cite{ming2022delving,miyai2023locoop} and achieving competitive results is by no means an easy task. Such observation can be certificated that in CLIP~\cite{radford2021learning}, CIFAR-100 does not obtain notably enhanced zero-shot performance against ImagenNet~($68.7$ v.s. $68.6$) although the latter contains richer categories and local texture information in images that is harder to classify. Furthermore, CLIP achieves superior results in ImageNet-100 with the same number of category~(Tab.~\ref{tab:imagenet100}) further validates the phenomenon. Moreover, the excessive training complexity and ineffectiveness make existing methods struggle to learn robust and large-scale representations from scratch, restraining their deployment in real-world scenarios. Conversely, we tackle the issue by proposing the effective prompt learning framework.

\noindent \textbf{Few-shot learning greatly improves efficiency with scarce samples.} Efficiency is one of the primary factors that are taken into consideration. To this end, we compare several critical aspects measuring efficiency, namely training epochs, number of shots and, most importantly, tuning time to attain extensive analysis of prompt learning framework for MisD. It can be seen in Fig.~\ref{fig:efficiency-analysis} that equipped with pre-trained vision language model, the tuning epoch of prompt learning is mere $\textbf{15\%}$~(30 v.s. 200) of the existing methods, indicating fast convergence and robust representation. Refraining from the utilization of the entire dataset for training, our method squeezes absolute tuning time to around $\textbf{87\%}$ using $\textbf{0.8\%}$~(4 v.s. 500 per category) of total training samples.
It strongly substantiates the superiority of few-shot learning framework and paves a promising way to enhance misclassification detection ability towards datasets much larger than traditional ones. Despite its efficiency, one potential drawback of our method would be larger GPU memory consumption since we employ the tremendous vision language model.

\begin{table}[t]
    \centering
    \caption{Comparison with conventional methods on CIFAR-10 and CIFAR-100. Our method achieves competitive results.}
    \vspace{5pt}
    \setlength{\tabcolsep}{5pt}
    \label{tab:cifar100}
        \resizebox{\linewidth}{!}{
            \begin{tabular}{l c c cccc cccc}
            \toprule[1.3pt]
            \multirow{2}{*}{\bftab{Method}} & \multirow{2}{*}{\bftab{Backbone}} & \multirow{2}{*}{\bftab{Setting}}&\multicolumn{4}{c}{\bftab{CIFAR-10}} &\multicolumn{4}{c}{\bftab{CIFAR-100}}\\
            \cmidrule(lr){4-7} \cmidrule(lr){8-11} 
             &&&\textbf{FPR95}$\downarrow$& \textbf{AURC}$\downarrow$ & \textbf{E-AURC}$\downarrow$ & \textbf{AUROC}$\uparrow$ & \textbf{FPR95}$\downarrow$& \textbf{AURC}$\downarrow$ & \textbf{E-AURC}$\downarrow$ & \textbf{AUROC}$\uparrow$ \\  
            \midrule
            Baseline\textcolor{gray}{[ICLR2017]]}~\cite{hendrycks2016baseline}&\multirow{5}{*}{\bftab{CNN}}&\multirow{5}{*}{\bftab{Full-tuning}}&32.65&5.94&4.74&92.77&61.60&60.16&33.88&86.98\\
            GradNorm\textcolor{gray}{[NeurIPS2021]}~\cite{huang2021importance}&&&94.15&43.20 &41.94&55.84  &84.76&150.02&119.43&66.62\\
            MaxLogit\textcolor{gray}{[ICML2022]}~\cite{hendrycks2022scaling}&&&53.60& 9.27& 6.50&89.01&58.78&81.28&50.71&83.25\\
            CRL\textcolor{gray}{[ICML2020]}~\cite{moon2020confidence}&&&38.92&6.23&4.39&\underline{93.46}&\underline{60.88}&62.44&32.66&\underline{87.79}\\
            FMFP\textcolor{gray}{[ECCV2022]}~\cite{zhu2022rethinking}&&&\underline{31.65}&\bftab{3.97}&\bftab{2.78}&\bftab{94.81}&61.03&\bftab{56.62}&\bftab{29.30}&\bftab{88.32}\\
            \midrule
            \rowcolor{gray!20} \textbf{Ours}&\bftab{VLM}&\bftab{Few-Shot}&\bftab{29.06}&\underline{4.78}&\underline{3.62}&93.06&\bftab{59.61}&\underline{58.09}&\underline{32.54}&87.02\\
            
            \bottomrule[1.3pt]
        \end{tabular}}
        \vspace{-10pt}
\end{table}

\subsection{Out-of-distribution Results}
Out-of-distribution detection, which aims to detect outliers that are not seen by the detector from known categories~\cite{ma2024towards, cheng2023average, cheng2024breaking}, is a task closely connected with misclassification detection. Concretely, they both aim to validate the robustness of the detector in open environment setting. The difference is that they are to identify misclassified samples and outlier samples, respectively. We also evaluate four common out-of-distribution detection datasets, \ie iNaturalist~\cite{van2018inaturalist}, SUN~\cite{xiao2010sun}, Places~\cite{zhou2017places} and Texture~\cite{cimpoi2014describing}, using ImageNet as in-distribution dataset to investigate the generalization ability of our method. Tab.~\ref{tab:ood} reveals that typical prompt learning methods concentrate on improving accuracy through cross-entropy loss and do not necessarily boost the ability to identify outliers, \ie, out-of-distribution samples. By contrast, owing to the proposed textual guided negative sample and negative loss, our method pushes pseudo features away from category prompts and also contributes to out-of-distribution detection, \ie, 4.48\% lower on FPR95($\downarrow$) and 0.64\% higher on AUROC($\uparrow$), which exhibits the efficacy and generalizability of our approach.

\begin{table}[hb]
    \centering
    \caption{Out-of-distribution detection performance. In-distribution dataset is ImageNet. Lower FPR95 and higher AUROC represent better performance.}
    \vspace{5pt}
    \setlength{\tabcolsep}{4pt}
    \label{tab:ood}
    \resizebox{\linewidth}{!}{
    \begin{tabular}{@{}ccccccccccc@{}}
            \toprule[1.3pt]
            \multirow{2}{*}{\textbf{Method}}& \multicolumn{2}{c}{iNaturalist} & \multicolumn{2}{c}{SUN} & \multicolumn{2}{c}{Places} & \multicolumn{2}{c}{Texture} & \multicolumn{2}{c}{\textbf{Average}} \\
            \cmidrule(lr){2-3} \cmidrule(lr){4-5} \cmidrule(lr){6-7} \cmidrule(lr){8-9} \cmidrule(lr){10-11}
             & FPR95$\downarrow$ & AUROC$\uparrow$ & FPR95$\downarrow$ & AUROC$\uparrow$ & FPR95$\downarrow$ & AUROC$\uparrow$ & FPR95$\downarrow$ & AUROC$\uparrow$ & \textbf{FPR95}$\downarrow$ & \textbf{AUROC}$\uparrow$  \\ 
             \midrule
    Baseline~\cite{radford2021learning}&\bftab{32.86}&\bftab{94.18}&38.18&92.55&43.84&90.10&59.39&85.82&43.56&90.66\\
    CoOp~\cite{zhou2022learning}&53.19&88.81&38.05&91.87&43.41&89.55&47.89&88.40&45.64&89.66\\
    \rowcolor{gray!20}\textbf{Ours}&35.19&93.04&\bftab{33.74}&\bftab{92.73}&\bftab{40.82}&\bftab{90.36}&\bftab{46.59}&\bftab{89.06}&\bftab{39.08}&\bftab{91.30} \\
        \bottomrule[1.3pt]
    \end{tabular}}
    \vspace{-5pt}
\end{table}

\section{Conclusion}
In this paper, we focus on computational inefficiency in existing works and propose to exploit pre-trained VLMs to improve the efficiency of misclassification detection. We build a prompt learning framework for MisD termed \textbf{\modelname{}}, which makes it possible to efficiently transfer knowledge to downstream tasks with few images for each category, thereby extending MisD to large-scale datasets. We also introduce effective adaptive pseudo samples to enhance confidence estimation by pushing category prompts away from pseudo features. We conduct comprehensive experiments on large-scale datasets with various domain shift to
validate its efficiency. Significant and consistent improvement across different shots and evaluation metrics corroborates the effectiveness and generalization ability of our method.

\section*{Acknowledgement}
This work has been supported by National Science and Technology Major Project (2022ZD0116500), National Natural Science Foundation of China~(62222609, 62076236), and CAS Project for Young Scientists in Basic Research~(YSBR-083).

\bibliographystyle{unsrt}  
\bibliography{main}  

\end{document}